\documentclass[10pt,twocolumn,letterpaper]{article}

\usepackage{iccv}
\usepackage{times}
\usepackage{epsfig,subfigure,overpic}
\usepackage{graphicx}
\usepackage{amsmath}
\usepackage{amssymb}
\usepackage{multirow}
\usepackage{enumitem}
\usepackage{booktabs}
\newcommand{\tabincell}[2]{\begin{tabular}{@{}#1@{}}#2\end{tabular}}

\usepackage[breaklinks=true,bookmarks=false]{hyperref}

\iccvfinalcopy 

\def\iccvPaperID{2108} 

\begin{document}

\title{PKU-MMD: A Large Scale Benchmark for Continuous Multi-Modal Human Action Understanding}

\author{
Chunhui Liu \;\;  Yueyu Hu \;\;  Yanghao Li \;\; Sijie Song  \;\; Jiaying Liu
\thanks{\footnotesize{
This dataset and work is funded by Microsoft Research Asia (project ID FY17-RES-THEME-013).}}
\\
Institute of Computer Science and Technology, Peking University\\
{\tt\small \{liuchunhui, huyy, lyttonhao, ssj940920, liujiaying\}@pku.edu.cn}
}

\maketitle

\begin{abstract}
Despite the fact that many 3D human activity benchmarks being proposed, most existing action datasets focus on the action recognition tasks for the segmented videos. There is a lack of standard large-scale benchmarks, especially for current popular data-hungry deep learning based methods. In this paper, we introduce a new large scale benchmark (PKU-MMD) for continuous multi-modality 3D human action understanding and cover a wide range of complex human activities with well annotated information. PKU-MMD contains 1076 long video sequences in 51 action categories, performed by 66 subjects in three camera views. It contains almost 20,000 action instances and 5.4 million frames in total. Our dataset also provides multi-modality data sources, including RGB, depth, Infrared Radiation and Skeleton. With different modalities, we conduct extensive experiments on our dataset in terms of two scenarios and evaluate different methods by various metrics, including a new proposed evaluation protocol 2D-AP. We believe this large-scale dataset will benefit future researches on action detection for the community.
\end{abstract}

\section{Introduction}
The tremendous success of deep learning have made data-driven learning methods get ahead with surprisingly superior performance for many computer vision tasks.
Thus, several famous large scale datasets have been collected to boost the research in this area~\cite{RussakovskyImageNet,caba2015activitynet}. ActivityNet~\cite{caba2015activitynet} is a superior RGB video dataset gathered from Internet media like YouTube with well annotated label and boundaries.

Thanks to the prevalence of the affordable color-depth sensing cameras like Microsoft Kinect, and the capability to obtain depth data and the 3D skeleton of human body on the fly, 3D activity analysis has drawn great attentions. As an intrinsic high level representation, 3D skeleton is valuable and comprehensive for summarizing a series of human dynamics in the video, and thus benefits the more general action analysis. Besides succinctness and effectiveness, it has a significant advantage of great robustness to illumination, clustered background, and camera motion. However, as a kind of popular data modality, 3D action analysis suffers from the lack of large-scale benchmark datasets. To the best of our knowledge, existing 3D action benchmarks have limitations in two aspects.

\begin{figure}[t]
   \centering
   \includegraphics[width=1\linewidth]{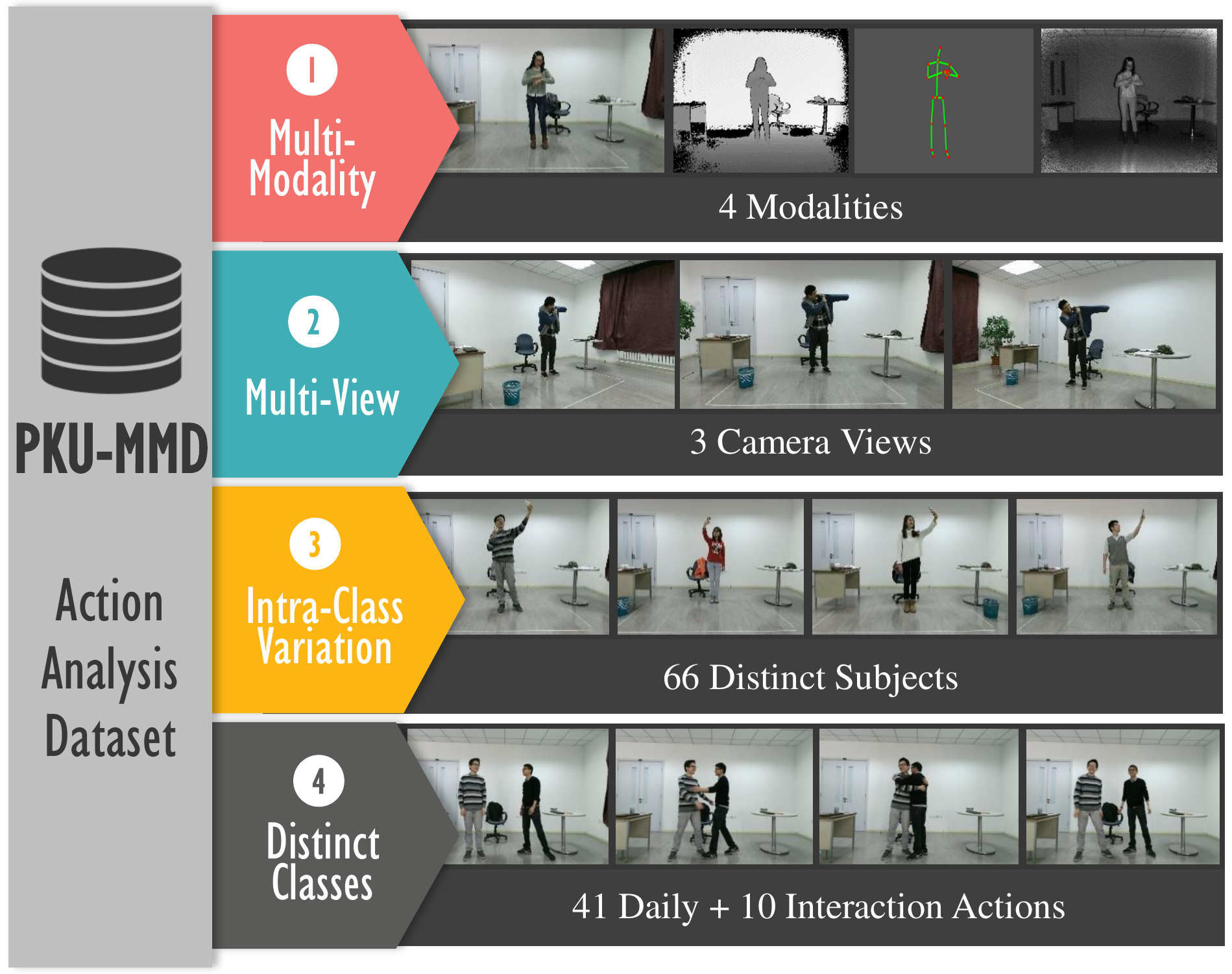}
   \caption{PKU Multi-Modalilty Dataset is a large-scale multi-modalities action detection dataset. This dataset contains 51 action categories, performed by 66 distinct subjects in 3 camera views.}
\label{fig:teaser}
\end{figure}

  $\bullet$ \textbf{Shortage in large action detection datasets:} Action detection plays an important role in video analytics and can be effectively studied through analysis and learning from massive samples. However, most existing 3D datasets mainly target at the task of action recognition for segmented videos. There is a lack of large scale multi-modal dataset for action detection. Additionally, previous detection benchmarks only contain a small number of actions in each video even in some large scale RGB datasets~\cite{caba2015activitynet}. There is no doubt that more actions within one untrimmed video will promote the robustness of action detection algorithms based on the sequential action modeling and featuring.

  $\bullet$ \textbf{Limitation in data modalities:} Different modalities (\emph{e.g.} optical flow, RGB, infrared radiation, and skeleton) intuitively capture features in different aspects and provide complementary information. For example, RGB frames can deliver appearance information but lack in motion representation, while optical flow is capable of describing motion but misses depth information which can be provided in skeleton. The combination of multi-modal data would benefit the application on action recognition and temporal localization. Traditional datasets focus mainly on one modality of action representation. Thus, it is worth exploiting multi-modal data with elegant algorithms for action analysis.

To overcome these limitations, we develop a new large scale continuous multi-modality 3D human activity dataset (PKU-MMD) for facilitating further study on human activity understanding, especially action detection. As shown in Figure~\ref{fig:teaser}, our dataset contains 1076 videos composed by 51 action categories, and each of the video contains more than twenty action instances performed by 66 subjects in 3 camera views. The total number of our dataset is 3,000 minutes and 5,400,000 frames. We provide four raw modalities: RGB frame, depth map, skeleton data, and infrared. More modalities can be further calculated such as optical flow and motion vector.

Besides, we propose a new 2D protocol to evaluate the precision-recall curve of each method in a much straightforward manner. Taking over-lapping ratio and detection confidence into account jointly, each algorithm can be evaluated with a single value, instead of a list of mean average precisions with corresponding overlap ratios. Several experiments are implemented to test both the capabilities of different approaches for action detection and the combination performance of different modalities.


\begin{table*}
\center
\caption{Comparison between different 3D datasets, including recognition and detection datasets.}
\vspace{0.5mm}
\begin{tabular}{c|c|c|c|c|c|c|c}
\hline
Datasets & Classes & Videos & \tabincell{c}{Labeled\\Instances} &\tabincell{c}{Actions\\per Video}& Modalities &\tabincell{c}{Temporal\\Localization} & Year \\
\hline
MSR-Action3D~\cite{li2010action}          & 20& 567 &567& 1 &D+Skeleton& No&2010\\
\hline
RGBD-HuDaAct~\cite{ni2013rgbd}         & 13& 1189&1189&1&RGB+D&No&2011\\
\hline
MSR-DailyActivity~\cite{wang2012mining}      & 16& 320 &320 & 1& RGB+D+Skeleton& No&2012\\
\hline
Act4~\cite{cheng2012human}             & 14& 6844 &6844 & 1 &RGB+D& No &2012\\
\hline
UTKinect-Action~\cite{xia2012view}     & 10& 200 & 200 & 1 &RGB+D+Skeleton& No &2012\\
\hline
3D Action Pairs~\cite{oreifej2013hon4d}      & 12& 360 &360 &1 & RGB+D+Skeleton & No&2013\\
\hline
DML-SmartAction~\cite{amiri2013non}       & 12& 932 &932 &1 & RGB+D & No&2013\\
\hline
MHAD~\cite{ofli2013berkeley}              & 11& 660 &660 &1 & RGB+D+Skeleton & No&2013\\
\hline
Multiview 3D Event~\cite{wei2013modeling} &8    &3815 &3815& 1 & RGB+D+Skeleton & No &2013\\
\hline
Northwestern-UCLA~\cite{wang2014cross}       &10 &1475 &1475 & 1 & RGB+D+Skeleton & No &2014\\
\hline
UWA3D Multiview~\cite{rahmani2014hopc}       &30 &$\sim$ 900 &$\sim$900& 1 & RGB+D+Skeleton & No &2014\\
\hline
Office Activity~\cite{wang20143d}      &20 &1180 &1180 & 1 & RGB+D & No &2014\\
\hline
UTD-MHAD~\cite{chen2015utd}            &27 & 861 & 861 & 1 & RGB+D+Skeleton & No &2015\\
\hline
TJU Dataset~\cite{liu2015coupled}         &22 & 1936 & 1936 & 1 & RGB+D+Skeleton & No &2015\\
\hline
UWA3D Multiview II~\cite{rahmani2016histogram}     &30 &1075 &1075 & 1 & RGB+D+Skeleton & No &2015\\
\hline
NTU RGB+D~\cite{shahroudy2016ntu}            &60 &56880  &56880 & 1 & RGB+D+IR+Skeleton & No &2016\\
\hline
\hline
G3D~\cite{bloom2012g3d}                & 20&  210& 1467& 7& RGB+D+Skeleton& Yes&2012\\
\hline
SBU Kinect interaction~\cite{yun2012two}  & 8 & 21 & 300&14.3& RGB+D+Skeleton& Yes&2012\\
\hline
CAD-120~\cite{sung2012unstructured}          & 20& 120 & $\sim$1200& $\sim$ 8.2 & RGB+D+Skeleton & Yes&2013\\
\hline
compostable Activities~\cite{lillo2014discriminative}    &16 &693 & 2529 & 3.6 & RGB+D+Skeleton& Yes &2014\\
\hline
Watch-n-Patch~\cite{wu2015watch}          &21 & 458 & $\sim$2500 & 2$\sim$7 & RGB+D+Skeleton & Yes &2015\\
\hline
OAD~\cite{li2016online}          &10 &59 & $\sim$700 &$\sim$12 & RGB+D+Skeleton & Yes &2016\\
\hline
\textbf{PKU-MMD}     &\textbf{51} &\textbf{1076} &\textbf{21545} & \textbf{20.02} & \textbf{RGB+D+IR+Skeleton} & \textbf{Yes} &\textbf{2017}\\
\hline
\end{tabular}
\label{ctable}
\vspace{-3mm}
\end{table*}

\section{Related Work}
\label{Related Work}
In this section, we briefly summarize the development of activity analysis. As a part of pattern recognition, activity analysis shows a common way of development in machine learning, where large scale benchmarks share familiar significance with magnificent methods. Here, we briefly introduce a series of benchmarks and approaches. For a more extensive conclusion of activity analysis we refer to corresponding survey papers~\cite{cai2016rgb,zhang2016rgb,aggarwal2014human,chen2013survey}.

\subsection{Development of Activity Analysis}
Early activity analysis mainly focuses on action recognition which consists of a classification task for segmented videos. Traditional methods mainly focus on hand-crafting features for video representation. Densely tracking points in the optical flow field with more features like Histogram of Oriented Gradient (HOG), Histogram of Flow (HOF) and Motion Boundary Histograms (MBH) encoded by Fisher Vector~\cite{perronnin2010improving, wang2013action} achieved a good performance. Recently, deep learning has been exploited for action recognition~\cite{simonyan2014two,wang2016temporal}. Deep approaches automatically learn robust feature representations directly from raw data and recognize actions synchronously with deep neural networks~\cite{szegedy2016inception}. To model temporal dynamics, Recurrent Neural Network (RNN) have also been exploited for action recognition. In~\cite{wu2015modeling,donahue2015long}, CNN layers are constructed to extract visual features while the followed recurrent layers are applied to handle temporal dynamics.

For action detection, existing methods mainly utilize either sliding-window scheme~\cite{wang2014action,sharaf2015real}, or action proposal approaches~\cite{wang2015cuhk}. These methods usually have low computational efficiency or unsatisfactory localization accuracy due to the overlapping design and unsupervised localization approach. Most methods are designed for offline action detection~\cite{tian2013spatiotemporal,sharaf2015real,wei2013concurrent}. However, in many new works, recognizing the actions on the fly before the completion of the action is well studied by a learning formulation based on a structural SVM~\cite{hoai2014max}, or a non-parametric moving pose framework~\cite{zanfir2013moving} and a dynamic integral bag-of-words approach~\cite{ryoo2011human}. LSTM is also used for online action detection and forecast which provides frame-wise class information. It forecasts the occurrence of start and end of actions.

As the fundamental requirement of research, videos source also determines the branches of action analysis. Early action analysis dataset mainly focuses on home surveillance activities like drinking or waving hands. The analysis of those simple indoor activities are the start of action recognition process. The advantages of this kind of videos lie in that they are usually easy and cheap to capture. However, collecting a large scale benchmark with cameras can be troublesome. Fortunately, the rapid development of Internet technology and data mining algorithms enable a new approach of collecting dataset from Internet third-way media like YouTube~\cite{soomro2012ucf101,caba2015activitynet}. As a result, RGB-based datasets achieve a grant level with hundreds of action labels and video sources in TB level. Recently, there are also several works focus on collecting different datasets of action type like TV-series~\cite{DeGeest2016}, Movies~\cite{laptev2008learning} and Olympic Games~\cite{karpathy2014large}.

With the launch of Microsoft Kinect, the diversity of action source becomes possible. Different input sources have been discussed such as Depth data and Skeleton data. Depth data provides a 3D information which is beneficial for action understanding. Skeleton, as a kind of high level representation of human body, can provide valuable and condensed information for recognizing actions. As Kinect devices provide a real-time algorithm to generate skeleton data from the information of RBG, depth, and infrared, skeleton becomes an ideal source to support real-time algorithm and to be transferred and utilized on some mobile devices like robots or telephones.

Despite of the diversity of source, action understanding still faces several problems, among which the top priority is the accuracy problem. Another problem is the poor performance of cross-data recognition. That is, existing approaches or machine learning models achieve good performances with training and test sets in similar environments conditions. Open domain action recognition and detection is still challenging.

\subsection{3D Activity Understanding Approaches}

For skeleton-based action recognition, many generative models have been proposed with superior performance. Those methods are designed to capture local features from the sequences and then to classify them by traditional classifiers like Support Vector Machine (SVM).
Those local features includes rotations and translations to represent geometric relationships of body parts in a Lie group~\cite{vemulapalli2014human,vemulapalli2016rolling}, or the covariance matrix to learn the co-occurrence of skeleton points~\cite{hussein2013human}. Additionally, Fourier Temporal Pyramids (FTP) or Dynamic Time Warping (DTW) are also employed to temporally align the sequences and to model temporal dynamics. Furthermore, many methods~\cite{goutsu2015motion, shahroudy2016multimodal} divide the human body into several parts and learn the co-occurrence information, respectively. A Moving Pose descriptor~\cite{zanfir2013moving} is proposed to mine key frames temporally via a k-NN approach in both pose and atomic motion features.

Most methods mentioned above focus on designing specific hand-crafted features and thus being limited in modeling temporal dynamics. Recently, deep learning methods are proposed to learn robust feature representations and to model the temporal dynamics without segmentation. In~\cite{du2015hierarchical}, a hierarchical RNN is utilized to model the temporal dynamics for skeleton based action recognition. Zhu \emph{et al.}~\cite{zhu2016co} proposed a deep LSTM network to model the inherent correlations among skeleton joints and the temporal dynamics in various actions. However, there are few approaches proposed for action detection on 3D skeleton data. Li \emph{et al.}~\cite{li2016online} introduced a Joint Classification Regression RNN to avoid sliding window design which demonstrates state-of-the-art performance for online action detection. In this work, we propose a large-scale detection benchmark to promote the study on continuous action understanding.

\subsection{3D Activity Datasets}
We have also surveyed other tens of well-designed action datasets which greatly improved the study of 3D action analysis. These datasets have promoted the construction of standardized protocols and evaluations of different approaches. Furthermore, they often provide some new directions in action recognition and detection previously unexplored. A comparison among several datasets and PKU-MMD is given in Table~\ref{ctable}.

\emph{MSR Action3D dataset}~\cite{li2010action} is one of the earliest datasets for 3D skeleton based activity analysis. This dataset is composed by instances chosen in the context of interacting with game consoles like \emph{high arm wave, horizontal arm wave, hammer, and hand catch}. The FPS is 15 frames per second and the skeleton data includes 3D locations of 20 joints.

\emph{G3D}~\cite{bloom2012g3d} is designed for real-time action recognition in gaming containing synchronized videos. As the earliest activity detection dataset, most sequences of \emph{G3D} contain multiple actions in a controlled indoor environment with a fixed camera, and a typical setup for gesture based gaming.

\emph{CAD-60}~\cite{sung2011human} \& \emph{CAD-120}~\cite{sung2012unstructured} are two special multi-modality datasets. Compared to \emph{CAD-60}, \emph{CAD-120} provides extra labels of temporal locations. However, the limited number of video instants is their downside.

\emph{ACT4}~\cite{cheng2012human} is a large dataset designed to facilitate practical applications in real life. The action categories in \emph{ACT4} mainly focus on the activities of daily livings. Its drawback is the limited modality.

\emph{Multiview 3D event}~\cite{wei2013modeling} and \emph{Northwestern-UCLA}~\cite{wang2014cross} datasets start to use multi-view method to capture the 3D videos. This method is widely utilized in many 3D datasets.

\emph{Watch-n-Patch}~\cite{wu2015watch} and \emph{Compostable Activities}~\cite{lillo2014discriminative} are the first datasets focusing on the continues sequences and the inner combination of activities in supervised or unsupervised methods. Those consist of moderate number of action instances. Also, the number of instance actions in one video is limited and thus cannot fulfill the basic requirement for deep network training.

\emph{NTU RGB+D}~\cite{shahroudy2016ntu} is a state-of-the-art large-scale benchmark for action recognition. It illustrates a series of standards and experience for large-scale data building. Recently reported results on this dataset have achieved agreeable accuracy on this benchmark.

\emph{OAD}~\cite{li2016online} dataset is a new dataset focusing on online action detection and forecast. 59 videos were captured by Kinect v2.0 devices which composed of daily activities. This dataset proposes a series of new protocols for 3D action detection and raises an online demand.

However, as the quick development of action analysis, these datasets are not able to satisfy the demand of data-driven algorithms. Therefore, we collect PKU-MMD dataset to overcome their drawbacks from the perspectives in Table~\ref{tab:perspetive}.

\begin{table}[htbp]
\small
   \centering
   \caption{The desirable properties of PKU-MMD dataset.}
   \begin{tabular}{l|l}

      \hline
      \textbf{Properties} & \textbf{Features} \\
      \hline
      \multirow{2}[2]{*}{Large Scale}
        & Extensive action categories. \\
      & Massive samples for each class. \\
      \hline
      \multirow{3}[2]{*}{Diverse Modality}
        & Three camera views. \\
        & Sufficient subject categories. \\
      & Multi-modality (RGB, depth, IR, \emph{etc.}). \\
      \hline
      \multirow{2}[2]{*}{Wide Application}
        & Continuous videos for detection. \\
      & Inner analysis of context-related actions. \\
      \hline
   \end{tabular}%
   \label{tab:perspetive}%
\end{table}%
\section{The Dataset}
\label{dataset}
\subsection{PKU-MMD Dataset}
PKU-MMD is our new large-scale dataset focusing on long continuous sequences action detection and multi-modality action analysis. The dataset is captured via the Kinect v2 sensor, which can collect color images, depth images, infrared sequences and human skeleton joints synchronously. We collect 1000+ long action sequences, each of which lasts about 3$\sim$4 minutes (recording ratio set to 30 FPS) and contains approximately 20 action instances. The total scale of our dataset is 5,312,580 frames of 3,000 minutes with 20,000+ temporally localized actions.

We choose 51 action classes in total, which are divided into two parts: 41 daily actions (drinking, waving hand, putting on the glassed, \emph{etc.}) and 10 interaction actions (hugging, shaking hands, \emph{etc.}).

We invite 66 distinct subjects for our data collection. Each subjects takes part in 4 daily action videos and 2 interactive action videos. The ages of the subjects are between 18 and 40. We also assign a consistent ID number over the entire dataset in a similar way in~\cite{shahroudy2016ntu}.

To improve the sequential continuity of long action sequences, the daily actions are designed in a weak connection mode. For example, we design an action sequence of \emph{taking off shirt, taking off cat, drinking water}and \emph{sitting down} to describes the scene that occur after going back home. Note that our videos only contain one part of the actions, either daily actions or interaction actions. We design 54 sequences and divide subjects into 9 groups, and each groups randomly choose 6 sequences to perform.

For the multi-modality research, we provide 5 categories of resources: depth maps, RGB images, skeleton joints, infrared sequences, and RGB videos. Depth maps are sequences of two dimensional depth values in millimeters. To maintain all the information, we apply lossless compression for each individual frame. The resolution of each depth frame is $512\times424$. Joint information consists of 3-dimensional locations of 25 major body joints for detected and tracked human bodies in the scene. We further provide the confidence of each joints point as appendix. RGB videos are recorded in the provided resolution of $1920\times1080$. Infrared sequences are also collected and stored frame by frame in $512\times424$.

\subsection{Developing the Dataset}
Building a large scale dataset for computer vision task is traditionally a difficult task. To collect untrimmed videos for detection task, the main time-consuming work is labeling the temporal boundaries. The goal of PKU-MMD is to provide a large-scale continuous multi-modality 3D action dataset, the items of which contain a series of compact actions. Thus we combine traditional recording approaches with our proposed validation methods to enhance the robustness of our dataset and improve the efficiency.

We now fully describe the collecting and labeling process for obtaining PKU-MMD dataset. Inspired by~\cite{shahroudy2016ntu}, we firstly capture long sequences from Kinect v2 sensors with a well-designed standards. Then, we rely on volunteers to localize the occurrences of dynamic and verify the temporal boundaries. Finally, we design a cross-validation system to obtain labeling correction confidence evaluation.

\begin{figure*}[t]
\vspace{1mm}
   \centering
    \subfigure[From top to bottom, these four rows show RGB, depth, skeleton and IR modalities, respectively.]
   {\includegraphics[width=1\linewidth]{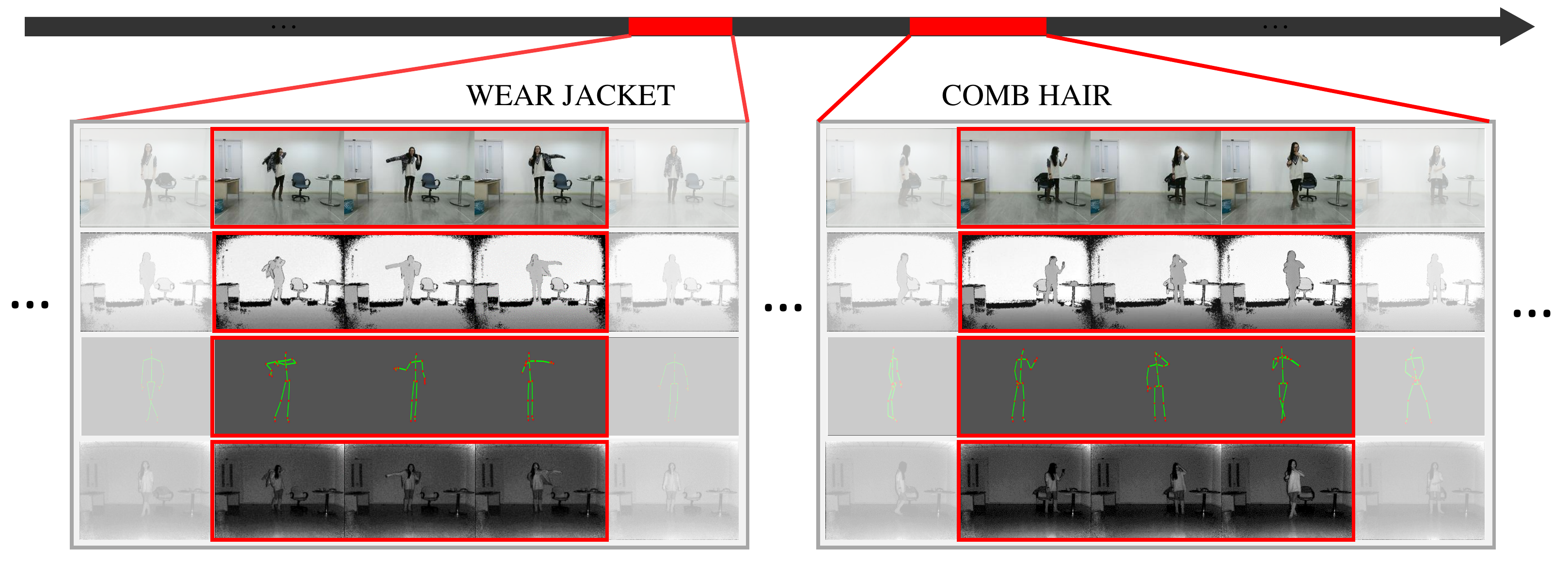}}
\vspace{1mm}
    \subfigure[We collect 51 actions performed by 66 subjects, including actions for single and pairs.]
   {\includegraphics[width=1\linewidth]{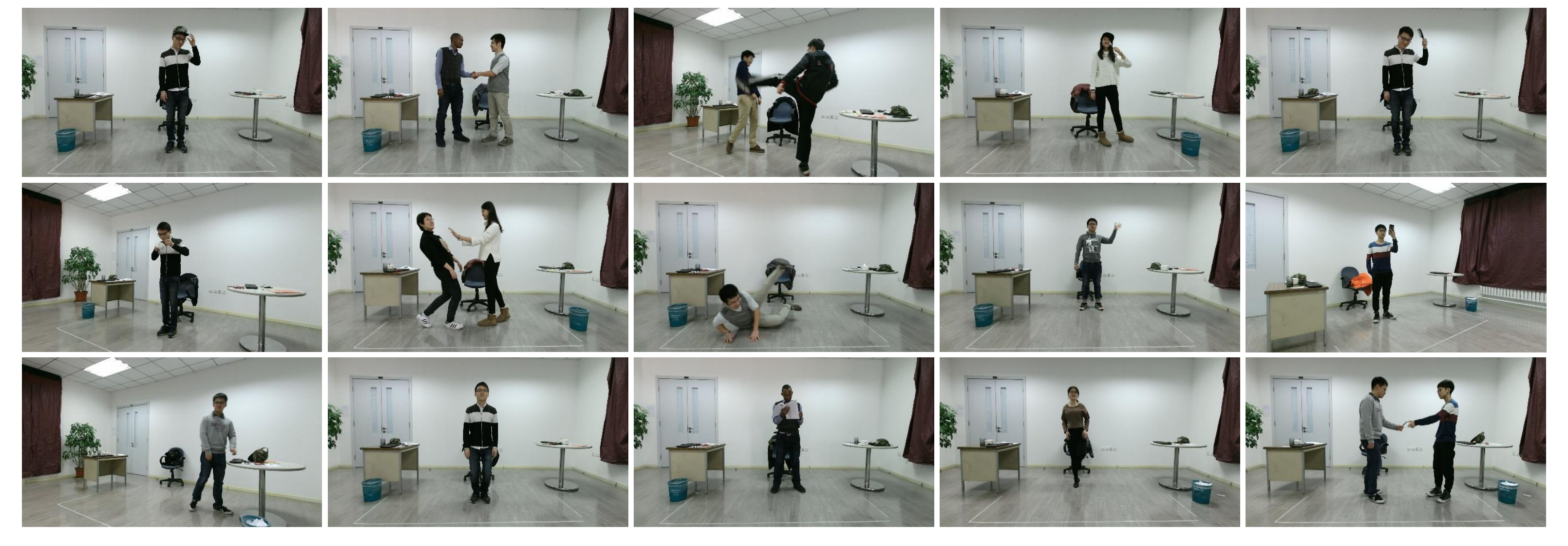}}

\vspace{1mm}
   \caption{Sample frames from PKU-MMD. The top figure shows an example of continuous action detection in multi-modality, and about 20 action instances can be found within one sequences. The bottom figure depicts the diversity in categories, subjects and camera viewpoints.}
\label{fig:overview}
\vspace{-3mm}
\end{figure*}

\textbf{Recording Multi-Modality Videos:} After designing several action sequences, we carefully choose a daily-life indoor environment to capture the video samples where some irrelevant variables are fully considered. Considering that the temperature changes will lead to the deviation of infrared sequences, we fully calculate the distance among the action occurrence, windows and Kinect devices. Windows are occluded for illumination consistency.
We use three cameras in the fixed angle and height at the same time to capture three different horizontal views. We set up an action area with $180cm$ as length and $120cm$ as width. Each subject will perform each action instances in a long sequence toward a random camera, and it is accepted to perform two continuous actions toward different cameras. The horizontal angles of each camera is $-45^{\circ}$, $0^{\circ}$, and $+45^{\circ}$, with a height of $120cm$. An example of our multi-modality data can be found in Figure~\ref{fig:overview}.

\textbf{Localizing Temporal Intervals:} At this stages, captured video sources are labeled on frame level. We employ volunteers to review each video and give the proposal temporal boundaries of each action presented in the long video. In order to keep high annotation quality, we merely employ proficient volunteers who have experiences in labeling temporal actions. Furthermore, there will be a deviation for the temporal labels of a same action from different persons. Thus we divide actions into several groups and the actions in each group are labeled by only one person. At the end of this process, we have a set of verified untrimmed videos that are associated to several action intervals and label correspondingly.

\textbf{Verifying and Enhancing Labels:}
Unlike recognition task which merely need one label for an trimmed video clip, the probability of error on temporal boundaries will be much higher. Moreover, during the labeling process we observe that approximate 10-frames expansion of action interval is sometimes accepted in some instance. To further improve the robustness of our dataset, we propose a system of labeling correction confidence evaluation to verify and enhance the manual labels. Firstly, we design basic evaluation protocol of each video, like \emph{If there is overlap of actions or Is the length of an action reasonable}. Thanks to multi-view capturing, we then use cross-view method to evaluate and verify the data label. The protocol guarantees the consistency of videos of each view.

\section{Evaluation Protocols}
\label{protocols}
To obtain a standard evaluation for the results on this benchmark, we define several criteria for the evaluation of the precision and recall scores in detection tasks. We propose two dataset partition settings with several precision protocols.

\subsection{Dataset Partition Setting}
This section introduces the basic dataset splitting settings for various evaluation, including cross-view and cross-subject settings.

\textbf{Cross-View Evaluation:} For cross-view evaluation, the videos sequences from the middle and right Kinect devices are chosen for training set and the left is for testing set. Cross-view evaluation aims to test the robustness in terms of transformation (\emph{e.g.}translation, rotation). For this evaluation, the training and testing sets have 717 and 359 video samples, respectively.

\textbf{Cross-Subject Evaluation:} In cross-subject evaluation, we split the subjects into training and testing groups which consists of 57 and 9 subjects respectively. For this evaluation, the training and testing sets have 944 and 132 long video samples, respectively. Cross-subject evaluation aims to test the ability to handle intra-class variations among different actors.

\subsection{Average Precision Protocols}
To evaluate the precision on the proposed action intervals with confidences, two tasks must be considered. One is to determine if the proposed interval is positive, and the other is to evaluate the performance of precision and recall. For the first task, there is a basic criterion to evaluate the overlapping ratio between the predicted action interval $I$ and the ground truth interval $I^{*}$ with a threshold $\theta$. The detection interval is correct when
\begin{equation}
\frac{|I \cap I^{*}|}{|I \cup I^{*}|} > \theta,
\end{equation}
where $I \cap I^{*}$ denotes the intersection of the predicted and ground truth intervals and $I \cup I^{*}$ denotes their union. So, with $\theta$, the $p(\theta)$ and $r(\theta)$ can be calculated.

\textbf{F1-Score:} With the above criterion to determine a correction detection, the F1-score is defined as
\begin{equation}
\mbox{F1}(\theta)=2 \cdot \frac{p(\theta) \times r(\theta)}{p(\theta) + r(\theta)}.
\end{equation}
F1-score is a basic evaluation criterion regardless of the information of the confidence of each interval.

\textbf{Interpolated Average Precision (AP):} Interpolated average precision is a famous evaluation score using the information of confidence for ranked retrieval results. With confidence changing, precision and recall values can be plotted to give a precision-recall curve. The interpolated precision $p_{interp}$ at a certain recall level $r$ is defined as the highest precision found for any recall level $r' \ge r$:
\begin{equation}
p_{interp}(r, \theta ) = \max_{r' \ge r} p(r',\theta).
\end{equation}
Note that $r$ is also determined by overlapping confidence $\theta$. The interpolated average precision is calculated by the arithmetic mean of the interpolated precision at each recall level.

\begin{equation}
\mbox{AP}(\theta) = \int_{0}^{1} p_{interp}(r, \theta) \; \mathrm{d}r.
\label{APFORM}
\end{equation}

\textbf{Mean Average Precision (mAP)}: With several parts of retrieval set $Q$, each part $q_j \in Q$ proposes $m_j$ action occurrences $\{d_1, \ldots d_{m_j}\}$ and $r_{jk}$ is the recall result of ranked $k$ retrieval results, then mAP is formulated by
\begin{equation}
\mbox{mAP}(\theta) = \frac{1}{\vert Q\vert} \sum_{j=1}^{\vert Q\vert} \frac{1}{m_j}
\sum_{k=1}^{m_j} p_{interp}(r_{jk},\theta).
\end{equation}
Note that with several parts of retrieval set $Q$, the AP score (\ref{APFORM}) is discretely formulated.

We design two splitting protocols: mean average precision of different actions (mAP$_{a}$) and mean average precision of different videos (mAP$_{v}$).

\textbf{2D Interpolated Average Precision:}
Though several protocols have been designed for information retrieval, none of them takes the overlap ratio into consideration. We can find that each AP score and mAP score is associated to $\theta$. To further evaluate the performance of precisions of different overlap ratios, we now propose the 2D-AP score which takes both retrieval result and overlap ratio of detection into consideration:
\begin{equation}
\mbox{2D-AP} = \iint_{r\in[0,1], \theta\in[0,1]} p_{interp}(r,\theta) \; \mathrm{d}r\mathrm{d}\theta.
\label{2DAPFORM}
\end{equation}

\section{Experiments}
\label{experiment}
This section presents a series of evaluation of basic detection algorithms on our benchmark. Due to the fact that there is few implementation for 3D action detection, these evaluations also serve to illustrate the challenge activity detection is and call on new explorations.

\subsection{Experiment Setup}
In this part, we implement several detection approaches for the benchmarking scenarios for the comparison on PKU-MMD dataset. Because of limited approaches for detection task, our base-line methods are divided into two phases, one is video representations and the other is temporal localizing and category classifying.

\begin{table*}
\center
\vspace{1mm}
\begin{tabular}{r||c|c|c|c||c|c|c|c}
\hline
Partition Setting&\multicolumn{4}{c||}{Cross-subject} & \multicolumn{4}{c}{Cross-view} \\
\hline
  $\theta$ & 0.1 & 0.3 & 0.5 & 0.7  & 0.1 & 0.3 & 0.5 & 0.7 \\
\hline
Deep RGB (DR) &0.507 & 0.323 & 0.147 & 0.024          & 0.617 & 0.439 & 0.221 & 0.051 \\
\hline
Deep Optical Flow (DOF) &0.626 & 0.402 & 0.168 & 0.023     & 0.697 & 0.482 & 0.234 & 0.044\\
\hline
Raw Skeleton (RS)       &0.479 & 0.325 & 0.130 & 0.014     & 0.545 & 0.352 & 0.159 & 0.026  \\
\hline
Convolution Skeleton (CS) & 0.493 & 0.318 & 0.121 & 0.010 &0.549 & 0.366 & 0.163 & 0.026  \\
\hline
RS + DR     &0.574 & 0.406 & 0.162 & 0.018      &0.675 & 0.498 & 0.255 & 0.050  \\
\hline
RS + DOF &0.643 & 0.438 & 0.171 & 0.021     & 0.741 & 0.537 & 0.258 & 0.045  \\
\hline
RS + DR + DOF  &0.647 & \textbf{0.476} & \textbf{0.199} & \textbf{0.026}&\textbf{0.747} & 0.584 &\textbf{0.306} & 0.066  \\
\hline
CS + DR + DOF  &\textbf{0.649} & 0.471 & \textbf{0.199} & 0.025 & 0.742 & \textbf{0.585} & \textbf{0.306} & \textbf{0.067}  \\
\hline
\end{tabular}
\vspace{3mm}
\caption{Comparison of different representation (mAP$_{a}$). Overlapping ratio $\theta$ vary from 0.1 to 0.7. Each representation feature is classified by three stacked BLSTM networks. First four rows show the result of single modality, and the last four rows evaluate the combination. We ensemble multi-modalities by averaging the predicted confidences.}
\label{ans_multi}
\vspace{-5mm}
\end{table*}

\subsubsection{Multi-Modality Representation}

In order to capture visual patterns in multi-modality input, we construct a series of video representations.

\textbf{Raw Skeleton (RS):} Raw skeleton can be directly considered as a representation for they containing high-order location context.

\textbf{Convolution Skeleton (CS):} Convolution skeleton is a new skeleton representation approach which add the temporal difference into raw skeleton with skeleton normalization. This method is illustrated in~\cite{zanfir2013moving} and is proven to be simple but effective.

\textbf{Deep RGB (DR):} For RGB-based action recognition, traditional motion features like HOG, HOF, and MBH are proven effective encoded by Fisher Vector~\cite{wang2011action,wang2013action}. However, Temporal Segment Networks (TSN)~\cite{wang2016temporal} have greatly improved the accuracy on several RGB-based dataset \emph{i.e.} UCF101, ActivityNet. In practice, we adopt features derived from convolution networks of TSN network that have been trained for action recognition as a robust RGB-based deep network feature.

\textbf{Deep Optical Flow (DOF):} Optical flow is well used in event detection, as it obtains a representation of motion dynamics. We fine-tune a deep BN-Inception network to learn the high-order features for temporal and spatial dynamics. This is motivated by the versatility and robustness of optical flow based deep features which are favorable in many recognition studies.


\subsubsection{Temporal Detection Method}
Here we introduce several approaches for action detection.

\textbf{Sliding Window + BLSTM/SVM:} Leveraging the insight from the RGB-based activity detection approaches, we design several slide-window detection approaches. For the classifier, we choose three stacked bidirectional LSTM (BLSTM) network and SVM motivated by the effectiveness of LSTM models~\cite{zhu2016co} and the agility of SVM classifier.

\textbf{Sliding Window + STA-LSTM:} Spatial-temporal attention network~\cite{song2016end} is a state-of-the-art work proposed for action recognition with unidirectional LSTM. It proposes a regularized cross-entropy loss to drive the model learning process which conducts automatic mining of discriminative joints together with explicitly learning and allocating the content-dependent attentions to the output of each frame to boost recognition performance.

\textbf{Joint Classification Regression RNN (JCRRNN):} Besides proposing the online action detection task, Li \emph{et al.}~\cite{li2016online} proposed a Joint Classification Regression RNN which implement frame level real-time action detection.

\begin{table}
\fontsize{8pt}{9pt}\selectfont\
\center
\vspace{1mm}
\begin{tabular}{r||c|c|c|c|c|c}
\hline
\multirow{2}{*}{Method} &\multicolumn{6}{c}{Cross-view} \\
\cline{2-7}
    &$\theta$ &F1 & AP & mAP$_{a}$& mAP$_{v}$ &2D-AP \\
\hline
\multirow{2}{*}{JCRRNN}
& 0.1 & 0.671 & \textbf{0.728} & \textbf{0.699} & \textbf{0.642} & \multirow{2}{*}{\textbf{0.460}}\\
\cline{2-6}
& 0.5 &\textbf{0.526} & \textbf{0.544} & \textbf{0.533} & \textbf{0.473} &\\
\hline
\multirow{2}{*}{SVM}
& 0.1 & 0.399 & 0.236 & 0.240 & 0.194 & \multirow{2}{*}{0.073}\\
\cline{2-6}
& 0.5 &0.131 & 0.031 & 0.036 & 0.031 & \\
\hline
\multirow{2}{*}{BLSTM}
& 0.1 &\textbf{0.676} & 0.525 & 0.545 & 0.508 & \multirow{2}{*}{0.187}\\
\cline{2-6}
& 0.5 &0.333 & 0.124 & 0.159 & 0.139 & \\
\hline
\multirow{2}{*}{STA-LSTM }
& 0.1 &0.613 & 0.468 & 0.476 & 0.439 & \multirow{2}{*}{0.180}\\
\cline{2-6}
& 0.5 &0.316 & 0.130 & 0.155 & 0.134 & \\
\hline
\hline
\multirow{2}{*}{Method} &\multicolumn{6}{c}{Cross-subject} \\
\cline{2-7}
    &$\theta$ &F1 & AP & mAP$_{a}$& mAP$_{v}$ &2D-AP \\
\hline
\multirow{2}{*}{JCRRNN}
& 0.1 & 0.500 & \textbf{0.479} & 0.452 & 0.431 & \multirow{2}{*}{\textbf{0.288}}\\
\cline{2-6}
& 0.5 &\textbf{0.366} & \textbf{0.339} & \textbf{0.325} & \textbf{0.297} &\\
\hline
\multirow{2}{*}{SVM}
& 0.1 &0.332 & 0.179 & 0.181 & 0.143 & \multirow{2}{*}{0.051}\\
\cline{2-6}
& 0.5 &0.092 & 0.016 & 0.021 & 0.018 &\\
\hline
\multirow{2}{*}{BLSTM}
& 0.1 &\textbf{0.629} & 0.464 & \textbf{0.479} & \textbf{0.442} & \multirow{2}{*}{0.164}\\
\cline{2-6}
& 0.5 &0.291 & 0.095 & 0.130 & 0.108 &\\
\hline
\multirow{2}{*}{STA-LSTM }
& 0.1 & 0.586 & 0.427 & 0.444 & 0.405 & \multirow{2}{*}{0.156}\\
\cline{2-6}
& 0.5 &0.284 & 0.101 & 0.131 & 0.116 &\\
\hline
\end{tabular}
\vspace{1mm}
\caption{Comparison of results among several approaches on 3D action detection with various metrics.}
\label{ans_ske}
\vspace{-5mm}
\end{table}

\subsection{PKU-MMD Detection Benchmarks}
In the detection task, the goal is to find and recognize all activity instances in an untrimmed video. Detection algorithms should provide the start and end points with action labels. We exploit the location annotations of PKU-MMD to compare the performances of above methods.

\subsubsection{Skeleton Based Scenarios}
As the skeleton is an effective representation, we implement several experiments to evaluate the ability to model dynamics and activity boundaries localizing. Table \ref{ans_ske} shows the comparison of different combination of skeleton representation and temporal featuring methods. It can be seen that the Deep Optical Flow beats other traditional features owning to its higher accuracy in motion description. STA-LSTM performs worse than BLSTM mainly due to the large margin in amount of parameters. Joint classification regression RNN achieves remarkable results, because it utilizes frame-level predictions and thus is more compatible with stricter localization requirements.

\begin{figure}[t]
\centering

\subfigure[Stride = 30]{
\includegraphics[width=0.47\linewidth]{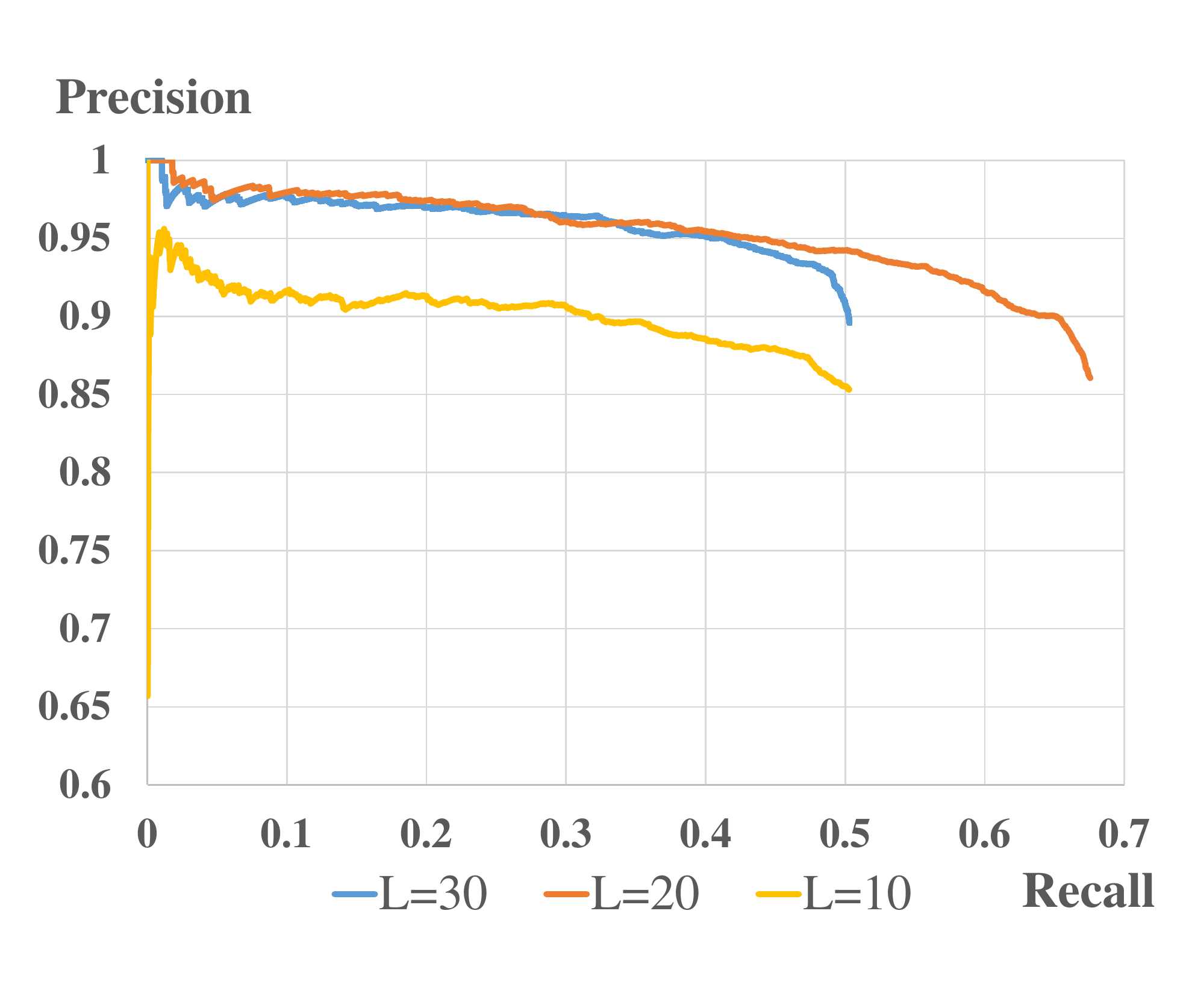}
}
\subfigure[Stride = 20]{
\includegraphics[width=0.47\linewidth]{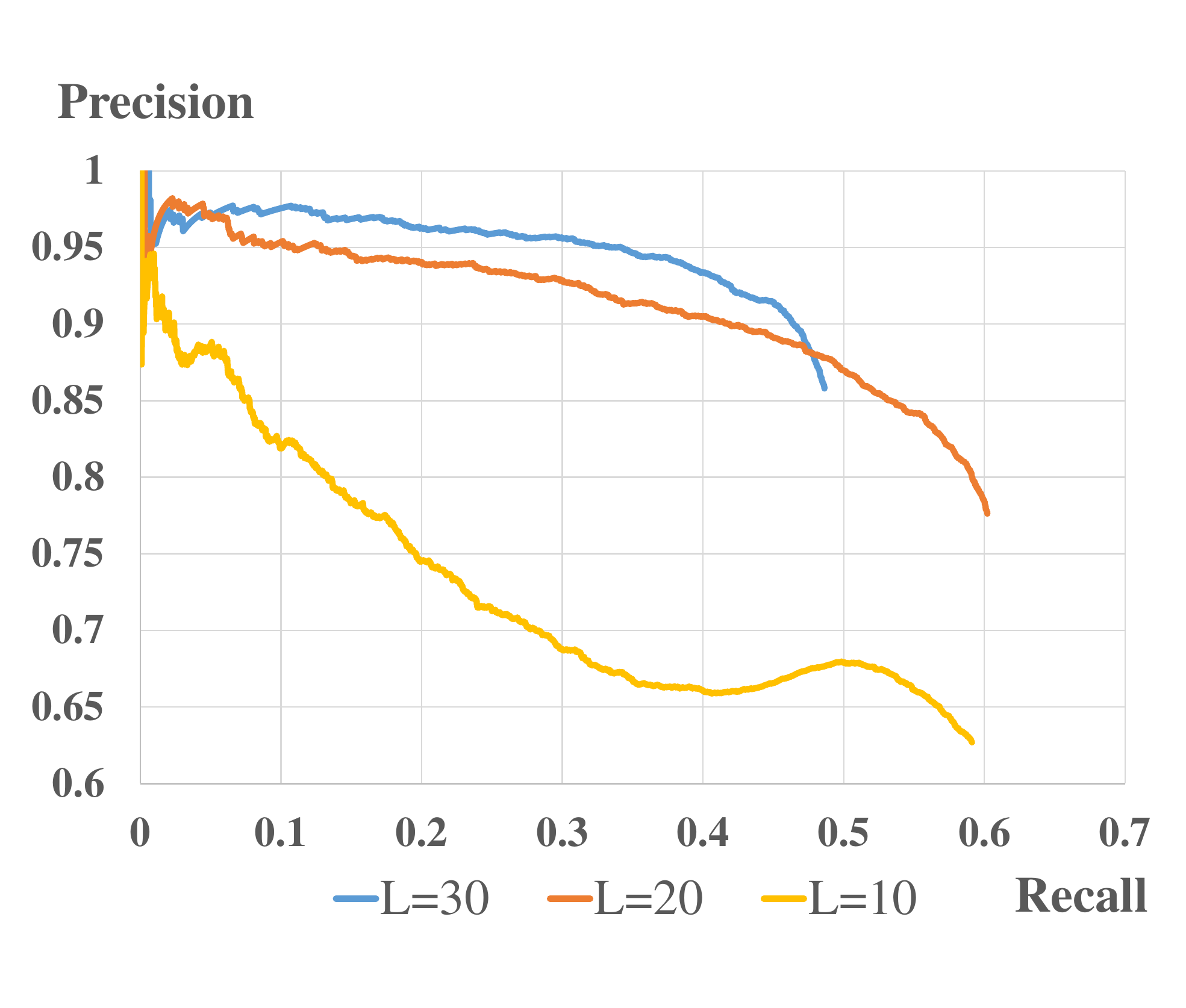}
}
\caption{Different Precision-Recall curves (overlapping ratio $\theta$ is set to 0.2) under different settings with different window size and stride. $L$ stands for the length of sliding windows.}
\label{PC_curve}
\vspace{-5mm}
\end{figure}
We further analyze the different performances with several sliding-window approaches. We show Precision-Recall curves of \emph{RS + BLSTM} method in Figure~\ref{PC_curve}. The performance is influenced by window size and stride. When stride is fixed, windows in smaller size contain less context information while noises can be involved by larger window size. However, smaller window size always leads to higher computation complexity. And obviously, smaller stride achieves better results due to dense sampling while costing more time. In our following experiments, we set 30 as window size and stride as a trade-off between performance and speed.

\vspace{-3mm}
\subsubsection{Multi-Modality Scenarios}
In this task, we evaluate the capability of detecting activity in multi-modality scenarios. Together with raw skeleton data, we calculate the first order differential of sequential skeleton input. Moreover, two deep features are extracted from fine-tuned deep convolution network: Deep RGB feature, Deep Optical Flow feature. We first evaluate the independent performances of different feature representations on long video sequences by classifying the 10-frame length sliding windows using SVM. Cross-view evaluation and cross-subject evaluation are both implemented in this evaluation. Then the combination of different modalities is observed, the result is shown in Table \ref{ans_multi}.

\section{Conclusion}
\label{conclusion}
In this paper, we propose a large-scale multi-modality 3D dataset (PKU-MMD) for human activity understanding, especially for action detection which demands localizing temporal boundaries and recognizing activity category. Performed by 66 actors, our dataset includes 1076 long video sequences, each of which contains 20 action instances of 51 action classes. Compared with current 3D datasets for temporal detection, our dataset is much larger (3000 minutes and 5.4 million frames in total) and contains much varieties (3 views, 66 subjects) in different aspects. The multi-modality attribution and larger scale of the collected data enable further experiments on deep networks like LSTM or CNN. Based on several detection retrieval protocols, we design a new 2D-AP evaluation for action detection task which takes both overlapping and detection confidence into consideration. We also design plenty experiments to evaluate several detection methods on PKU-MMD benchmarks. The results show that existing methods are not satisfied in terms of performance. Thus, large-scale 3D action detection is far from being solved and we hope this dataset can draw more studies in action detection methodologies to boost the action detection technology.

\pagebreak

{\small
\bibliographystyle{ieee}
\bibliography{iccv_refs}
}
\end{document}